\documentclass[10pt,twocolumn,letterpaper]{article}

\usepackage{cvpr}              
\definecolor{cvprblue}{rgb}{0.21,0.49,0.74}
\usepackage[pagebackref,breaklinks,colorlinks,allcolors=cvprblue]{hyperref}

\def\paperID{*****} 
\def\confName{CVPR}
\def\confYear{2026}

\title{Advancing Complex Video Object Segmentation via Tracking-Enhanced Prompt: The 1st Winner for 5th PVUW MOSE Challenge}

\author{
Jinrong Zhang$^{1}$, Canyang Wu$^{1}$, Xusheng He$^{1}$, Weili Guan$^{1,2}$, Jianlong Wu$^{1,2}$, Liqiang Nie$^{1,2}$
\\ 
$^1$Harbin Institute of Technology, Shenzhen, China $^2$Shenzhen Loop Area Institute, China \\ Team: HITsz\_Dragon
}

\begin{document}
\maketitle
\begin{abstract}
In the Complex Video Object Segmentation task, researchers are required to track and segment specific targets within cluttered environments, which rigorously tests a method's capability for target comprehension and environmental adaptability. Although SAM3, the current state-of-the-art solution, exhibits unparalleled segmentation performance and robustness on conventional targets, it underperforms on tiny and semantic-dominated objects. The root cause of this limitation lies in SAM3's insufficient comprehension of these specific target types. To address this issue, we propose \textbf{TEP}: Advancing Complex Video Object Segmentation via \textbf{T}racking-\textbf{E}nhanced \textbf{P}rompts. As a training-free approach, TEP leverages external tracking models and Multimodal Large Language Models to introduce tracking-enhanced prompts, thereby alleviating the difficulty SAM3 faces in understanding these challenging targets. Our method achieved first place (56.91\%) on the test set of the PVUW Challenge 2026: Complex Video Object Segmentation Track.
\end{abstract}    
\section{Introduction}
\label{sec:intro}

Video Object Segmentation (VOS)\cite{Tian_2025_CVPR} fundamentally aims to densely track and segment specific target objects across video sequences given their initial state. While conventional VOS has witnessed remarkable progress, it often assumes relatively distinct objects and clean backgrounds. Pixel-level and frame-level understanding tasks in videos have long been a major focus of researchers\cite{zhang2025end,11210093,wu2025survey}. As research progresses, researchers have continuously designed increasingly complex tasks to evaluate and verify model performance. Complex Video Object Segmentation\cite{zhang2025sec} presents a far more demanding challenge by placing targets within highly cluttered, occluded, or dynamic environments. Compared to standard VOS, this complex variant rigorously evaluates a model's intrinsic capability for target comprehension and its adaptability against severe environmental interference, pushing the boundaries of what current perception systems can achieve.

\begin{figure}[t]
    \centering
    \includegraphics[width=\linewidth]{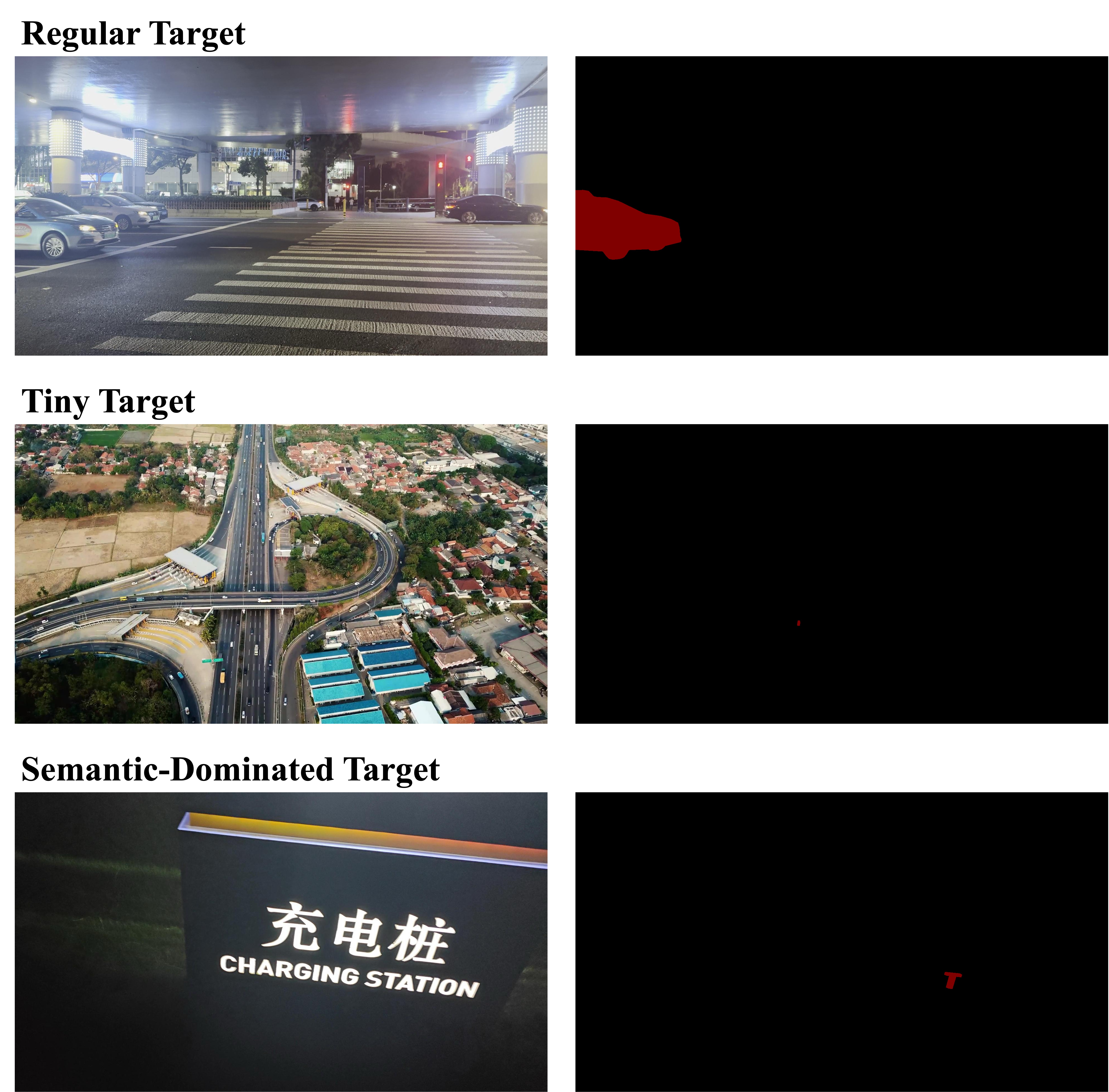}
    \caption{Visualization of the three target types in the MOSE v2 dataset, highlighting the challenges posed by tiny targets and semantic-dominated targets in complex environments. Tiny targets are visually minuscule and easily overwhelmed by background interference, while semantic-dominated targets possess highly similar visual features that can be easily confused with intra-class instances.}
    \label{fig:mose_data_v2}
\end{figure}

\begin{figure*}[t]
    \centering
    \includegraphics[width=\linewidth]{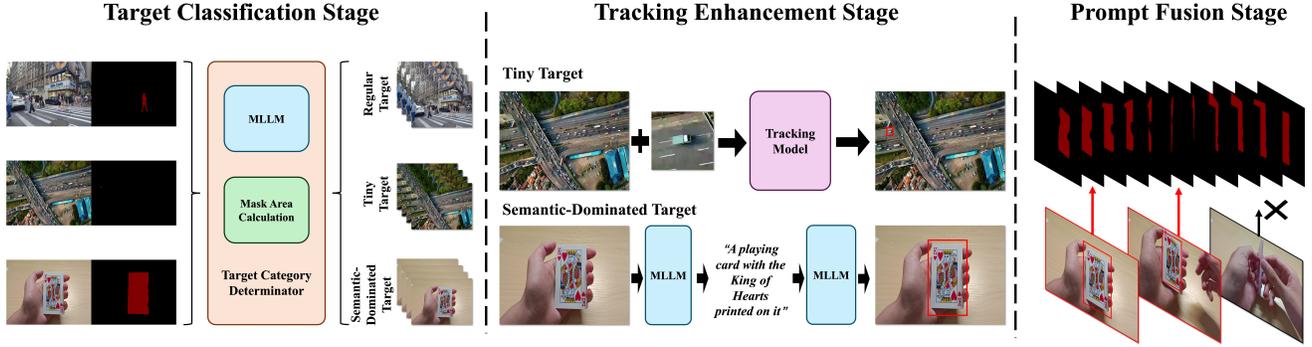}
    \caption{The architecture of the proposed TEP framework, which consists of three main stages: Target Classification, Tracking Enhancement, and Prompt Fusion. The Target Classification stage categorizes video targets into Regular, Tiny, and Semantic-Dominated types using mask area calculations and MLLM. The Tracking Enhancement stage employs tracking methods for Tiny Targets and MLLM for Semantic-Dominated Targets to generate bounding box prompts. The Prompt Fusion stage dynamically integrates these prompts into the SAM3 pipeline based on IoU and confidence scores to enhance segmentation accuracy and stability.}
    \label{fig:TEG}
\end{figure*}

Recent research in Complex VOS has predominantly focused on enhancing spatiotemporal feature matching and memory mechanisms\cite{Videnovic_2025_CVPR}. Methods such as XMem\cite{cheng2022xmem} and Cutie\cite{cheng2024putting} have significantly improved long-term tracking by designing sophisticated memory architectures to associate targets robustly across frames. Building upon the success of foundation models, SAM3\cite{carion2025sam} has emerged as the current state-of-the-art (SOTA) solution in this domain. SAM3 exhibits unparalleled segmentation performance and exceptional robustness when dealing with conventional targets, demonstrating an impressive ability to delineate sharp boundaries. However, despite its dominance in standard scenarios, its performance begins to falter under certain extreme conditions.

A critical limitation of the prevailing research trajectory is its over-reliance on refining memory management. While these approaches excel at retrieving historical target appearances, they fundamentally lack the mechanism to optimize the model's deeper, semantic comprehension of the targets themselves or to proactively adapt to highly chaotic scenes. This deficiency remains a bottleneck even for advanced foundation models like SAM3. When confronted with complex scenarios, SAM3 inherently struggles to maintain a deep understanding of the targets, making it susceptible to tracking drift and fragmentation.

To overcome these intrinsic limitations, we propose TEP: advancing complex video object segmentation via tracking-enhanced prompts. Traditional VOS architectures, constrained by their model scale, struggle to independently develop the deep target understanding required for complex scenarios, a capability that is naturally possessed by specialized tracking models and Multimodal Large Language Models (MLLMs). Specifically, dedicated tracking models excel at localizing tiny, inconspicuous objects, while MLLMs possess a profound capacity for interpreting semantic-dominated targets. In our TEP framework, we first employ an MLLM to categorize the video context into three distinct target types: regular, tiny, and semantic-dominated. Regular targets are processed directly by the robust SAM3 baseline. For tiny and semantic-dominated targets, we dynamically introduce a tracking-enhanced prompt, provided by an external tracking model and an MLLM, respectively. These prompts serve as crucial auxiliary guidance, significantly lowering the cognitive burden on SAM3 when processing intermediate frames.

By integrating this targeted guidance, the proposed TEP framework effectively enhances SAM3's tracking capabilities for special and challenging objects. Furthermore, TEP boasts exceptional transferability and modularity. When encountering extremely unique target categories, our approach circumvents the computationally prohibitive need to fine-tune the entire SAM3 foundation model. Instead, one only needs to train a lightweight, specialized tracking model to provide the necessary prompt enhancement, making the entire pipeline highly efficient and easy to reproduce. Ultimately, TEP achieved 1st place in the MOSE Track of the 5th PVUW Challenge 2026, demonstrating its superior performance and unwavering robustness in the most complex visual scenarios.
\section{TEP Solution}
\label{sec:solution}

\subsection{Overall Architecture}

The proposed Tracking-Enhanced Prompt (TEP) framework aims to mitigate the inherent limitations of foundation models in highly cluttered environments by dynamically injecting targeted prior knowledge. As shown in Figure \ref{fig:TEG}, the TEP pipeline operates through three sequential stages: Target Classification, Tracking Enhancement, and Prompt Fusion. Initially, a coarse classification mechanism filters video sequences containing special category targets. These selected sequences are then routed to the Tracking Enhancement stage, which generates robust, category-specific bounding box prompts. Finally, the Prompt Fusion stage dynamically integrates these auxiliary prompts into the base segmentation model (SAM3) to ensure temporal stability and boost overall segmentation accuracy.

\begin{table*}[t]
  \caption{Quantitative comparison of our TEP framework against the SAM3 baseline on the MOSE v2 test set. The metrics include overall J\&F, J, F, and specific evaluations for disappearance and reappearance scenarios.}
  \centering
  \label{tab:quantitative_results}
\begin{tabular}{c|ccccccc}
\hline
Method                       & $J\&\dot{F}$ & J     & $\dot{F}$ & $J$\&$\dot{F}_{disappear}$ & $J$\&$\dot{F}_{reappear}$ & F     & J\&F  \\ \hline
SAM3                         & 46.63     & 44.76 & 48.50  & 56.94                & 31.54               & 51.04 & 47.90 \\
TEP (MLLM)         & 55.57     & 53.44 & 57.70  & 61.36                & 42.45               & 60.75 & 57.10 \\
TEP (Tracking Model \& MLLM) & 56.91     & 54.71 & 59.12  & 62.09                & 44.11               & 62.39 & 58.55 \\ \hline
\end{tabular}
\end{table*}

\subsection{Target Classification Stage}

Because complex video sequences exhibit diverse challenges, uniformly processing all targets is suboptimal. In this initial stage, we employ an evaluation metric based on mask area calculations coupled with a Multimodal Large Language Model (MLLM) to categorize targets into three distinct types:

\noindent\textbf{Regular Targets}: Objects with a standard mask area whose visual appearance lacks distinct, easily verbalized semantic features that would separate them from intra-class instances in the same video. These targets are processed directly and entirely by the SAM3 baseline.

\noindent\textbf{Tiny Targets}: Objects characterized by an exceptionally small initial mask area, making them highly susceptible to feature degradation over time.

\noindent\textbf{Semantic-Dominated Targets}: Objects with a standard mask area that possess highly distinct visual attributes (e.g., a specific logo, unique text, or distinctive clothing) that can be unambiguously described using natural language to distinguish them from surrounding similar instances.

Videos containing Tiny and Semantic-Dominated targets are forwarded to the subsequent Tracking Enhancement stage.

\subsection{Tracking Enhancement Stage}

To address the specific vulnerabilities identified in the classification stage, we deploy two distinct, specialized tracking paradigms to generate reliable bounding box (bbox) coordinates for intermediate frames:

\noindent\textbf{Handling Tiny Targets via Tracking Model}: Traditional tracking methods often rely on pixel-level feature diffusion, which frequently leads to severe tracking drift when tiny targets undergo prolonged occlusion. To counter this, we employ SUTrack, an image-prompted tracking methodology. Using images as prompts for detection and tracking can effectively overcome the limitation when targets are difficult to describe with text. By utilizing the RGB image crop of the target as an explicit prompt, SUTrack forces the model to continuously reference and memorize the core visual features of the object, significantly reinforcing the tracking robustness for tiny entities\cite{zhang2025just,jiang2024t,oquab2023dinov2}.

\noindent\textbf{Handling Semantic-Dominated Targets via MLLMs}: In scenarios where the target is camouflaged among numerous homogeneous instances, conventional visual trackers often fail. However, these targets typically possess a unique feature that is easily articulated. Leveraging the superior vision-language alignment capabilities of MLLMs, we utilize Qwen3.5. The MLLM first generates a precise textual description of the target based on the reference frame. Subsequently, it performs frame-by-frame object detection guided by this textual prompt, successfully isolating the target from similar instances and outputting the corresponding bounding boxes.

\subsection{Prompt Fusion Stage}

The final stage governs the dynamic integration of the auxiliary bounding boxes acquired from the Tracking Enhancement stage into the SAM3 pipeline. During the processing of intermediate video frames, we calculate the Intersection over Union (IoU) between the bounding box of SAM3’s predicted mask and the auxiliary bbox provided by our enhancement module. If the IoU falls below a predefined threshold, indicating a potential tracking failure or drift by SAM3, a prompt-switching mechanism is triggered:

\noindent\textbf{For Tracking Model Bboxes}: We evaluate the confidence score of the generated bbox. If the confidence is excessively low, the system discards the auxiliary prompt and continues relying on SAM3’s inherent pixel diffusion. Conversely, if the confidence score is sufficiently high, the SUTrack bbox is injected as a new, corrective prompt to guide SAM3's subsequent segmentation.

\noindent\textbf{For MLLM-Generated Bboxes}: We introduce a MLLM as a judge mechanism. The MLLM takes the visual crops bounded by both SAM3’s predicted mask and the MLLM's detection bbox, and compares them against the original target in the first frame. The MLLM evaluates which crop preserves higher semantic and visual fidelity to the reference target. The most accurate bounding box is then selected as the definitive prompt to guide the ongoing VOS process.

\section{Experiments}

\subsection{Challenge Description and Dataset}

\begin{figure*}[t]
    \centering
    \includegraphics[width=\linewidth]{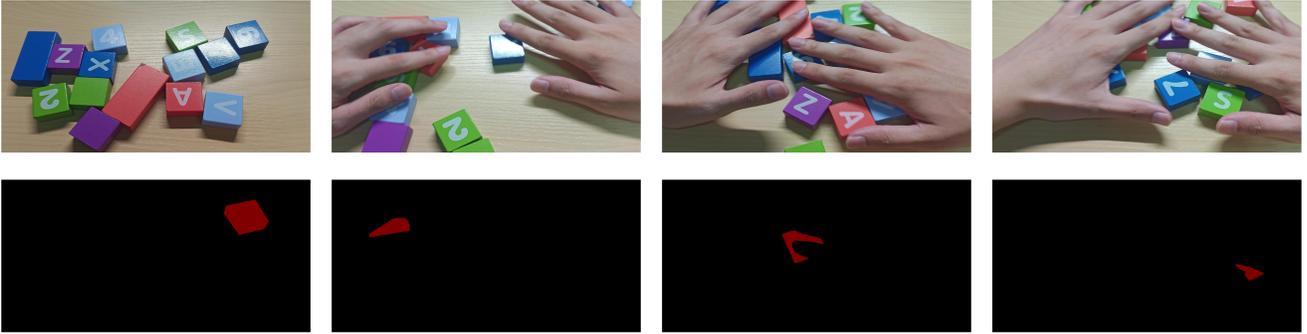}
    \caption{Qualitative visualization of segmentation results on the MOSE v2 test set. The visualizations demonstrate that the tracking and segmentation contours for both tiny and semantic-dominated targets remain clean, tight, and temporally stable, validating the efficacy of our prompt-enhancement strategy.}
    \label{fig:visualization_v2}
\end{figure*}

The MOSE Track of the 5th PVUW Challenge (2026) significantly raised the benchmark by upgrading the dataset from MOSE v1 to MOSE v2\cite{MOSEv2}. While MOSE v1\cite{MOSE} already emphasizes challenging cases such as object disappearance-reappearance, small objects, heavy occlusions, and crowded scenes, MOSE v2 introduces a vastly expanded scale (over 5,000 videos) and intensified real-world complexities. Notably, v2 incorporates severe environmental degradation, including adverse weather conditions and low-light scenes, which drastically increases the difficulty of temporal tracking. Amidst these exacerbated challenges, the vulnerability of current foundation models to tiny targets and semantic-dominated objects becomes particularly pronounced. These specific targets are highly susceptible to being lost in cluttered, visually degraded backgrounds or being confused with semantically similar intra-class instances, making them the primary bottlenecks for achieving high segmentation accuracy in the 2026 challenge.

\subsection{Implementation Details}

To directly tackle these category-specific bottlenecks, our proposed TEP adopts a highly modular and flexible design. In our specific implementation for the challenge, we employ SUTrack as the dedicated tracking model to handle tiny targets, while utilizing the powerful Qwen3.5-397B-A17B\cite{qwen35blog} as our chosen MLLM to process semantic-dominated targets. Benefiting from this decoupled architecture, TEP remains structurally straightforward yet highly adaptable. This modularity allows researchers to seamlessly plug in or swap out different tracking modules or MLLMs depending on the unique characteristics of the target categories they are investigating. Consequently, TEP can deliver robust, specialized performance enhancements precisely where they are needed, without the computational burden of retraining the foundational segmentation network.

\subsection{Results}

Our quantitative and qualitative results demonstrate the substantial superiority of the TEP framework in complex scenarios. As shown in Table \ref{tab:quantitative_results}, the baseline ``SAM3" (representing direct, unprompted output) struggles to maintain temporal consistency under the severe conditions of MOSE v2. Integrating MLLM, ``TEP (MLLM)", yields noticeable improvements specifically on semantic-dominated object samples. However, the complete pipeline, ``TEP (Tracking Model \& MLLM)", achieves the most significant performance leap by comprehensively addressing both tiny and semantic-dominated target vulnerabilities.

Most notably, our method secures a remarkable improvement on $J$\&$\dot{F}_{reappear}$. This metric is recognized as a demanding indicator in the evaluation suite, as it strictly tests a model's ability to successfully re-associate targets after prolonged periods of complete occlusion, a challenge that TEP addresses effectively. As shown in Figure \ref{fig:visualization_v2}, our qualitative visualizations corroborate these findings. The visualizations intuitively demonstrate that the tracking and segmentation contours for both tiny and semantic-dominated targets remain exceptionally clean, tight, and temporally stable, validating the efficacy of our prompt-enhancement strategy.

\section{Conclusion}

In this technical report, we present TEP: Advancing Complex Video Object Segmentation via Tracking-Enhanced Prompt, a novel framework designed to address the critical challenges posed by complex video scenarios. By strategically integrating specialized tracking models and Multimodal Large Language Models as auxiliary prompt generators, TEP effectively enhances the segmentation capabilities of the SAM3 foundation model, particularly for tiny and semantic-dominated targets. Our comprehensive evaluation on the MOSE v2 dataset demonstrated that TEP significantly outperforms the SAM3 baseline, achieving superior temporal stability and accuracy in highly cluttered environments. The modular design of TEP allows for flexible adaptation to various target categories, making it a robust and efficient solution for advancing the state-of-the-art in complex video object segmentation.
{
    \small
    \bibliographystyle{ieeenat_fullname}
    \bibliography{main}
}


\end{document}